# Automatic Construction of Chinese Verb Collostruction Database

Xuri TANG[1]    Daohuan LIU

Huazhong University of Science and Technology

## 1 Introduction

The phenomenal success of end-to-end neural-network based learning in natural language processing has side-lined research efforts in the construction of lexical knowledge database (Majewska & Korhonen, 2023). Nevertheless, researchers have noted several problematic aspects in large language models (LLMs), one representative of end-to-end neural-network models, such as hallucination (Ji et al., 2023) and lack of interpretability and controllability (Zhang, Song, Li, Zhou, & Song, 2023), which can be overcome via the unification of knowledge database and LLMs to support application in high-stakes scenarios such as medical diagnosis, legal judgment (Pan et al., 2024). Similar insights in the circle of computational lexical semantics have attracted more research on semi-automatic and automatic construction of lexical knowledge database, particularly verb knowledge database (Kawahara, Peterson, Popescu, & Palmer, 2014a; Majewska & Korhonen, 2023; Reichart & Korhonen, 2013; Roberts, 2022; Scarton et al., 2014; Sun & Korhonen, 2011). In line with the above paradigm, this paper reports a fully unsupervised approach to automatic construction of verb collostruction database for Chinese. It makes the following contributions to the field:

(1) A formal definition of verb collostruction as the basic unit of verb knowledge database that possesses the design features of functional independence and graded typicality with negative evidence;

(2) An unsupervised algorithm for automatic generation of verb collostructions on the basis of syntactic parsing, DBSCAN-based clustering, and word embeddings;

(3) An evaluation experiment of grammar error detection of Chinese verbs demonstrating advantages over a state of the art of LLM.

---

[1] Corresponding author. Please send suggestions and comments to the email address: xrtang@hust.edu.cn.

The paper is organized as follows. It first discusses the rationales for a verb collostruction database, presents the formal definition of verb collostruction and its design features, explicates the algorithm of automatic generation of verb collostructions, reports the evaluation of the constructed collostruction database with verb usage classification, and concludes with discussions on the generalization of the proposed algorithm to other languages.

## 2 Rationales for verb collostruction database

### 2.1 Lexical knowledge database vs. end-to-end neural-network based learning

Recent research has shown that the end-to-end neural-network based learning is flawed despite the seminal success obtained by LLMs and other applications. With LLMs, the criticism involves hallucination, lack of factual knowledge, and lack of interpretability, because they implicitly represent knowledge in their parameters and perform probabilistic and indeterministic inferences (Pan et al., 2024). Metaphysically, such a learning method falls into the category of inductive-statistical explanation (or prediction)[2] (Hempel, 1965, p. 333), with the neural network accounting

---

[2] Following (Hempel, 1965, p. 333), scientific explanation of linguistic phenomena may fall into three basic categories: deductive-nomological, deductive-statistical, and inductive-statistical. The schematic model of deductive-nomological is summarized as follows:

$$[(c_1, c_2, \ldots, c_k), (l_1, l_2, \ldots, l_r)] \Rightarrow E$$

An explanation $E$ is derived from a sequence of facts $(c_1, c_2, \ldots, c_k)$ and a sequence of general laws $(l_1, l_2, \ldots, l_r)$. Because of its essential reliance on laws and theoretical principles, deductive-nomological explanation may be expected to show a close affinity to scientific prediction and prepares the mind to understand and ascertain facts without need of recourse to experiments (Hempel, 1965, pp. 364-365). Deductive-statistical explanation amounts to the deductive subsumption of a narrower statistical uniformity under more comprehensive ones, or to calculate certain derivative probabilities on the basis of other probabilities which have been empirically ascertained or hypothetically assumed (Hempel, 1965, p. 380). The schematic model is denoted as follows:

$$p(F, G) = r, F \sqsubset G$$

That is, given G subsumes F, the statistical probability for an event of kind F to be also of kind G is r. The deductive nature of the above category of explanation originates from statistical laws accounting for G subsuming F.

Inductive-statistical explanation, nevertheless, assumes that F and G are two different events and provides an explanation using the following schematization:

$$if\ p(R, F \cdot G)\ is\ close\ to\ 1\ and\ F_i \cdot G_i \Rightarrow R_i, F \neq G$$

This category arrives at the explanation ($R_i$) because (1) the empirical probability $p(R, F \cdot G)$ is very high and $F_i$ co-occurs with $G_i$. One particular logical phenomenon with inductive-statistical explanation is ambiguity, i.e., for a proposed inductive-statistical explanation, "there will often exist a rival argument of the same probabilistic form and with equally true premises which confers near certainty upon the nonoccurrence of the same event" (Hempel, 1965, p. 395).

for the high co-occurring probability from one end to another. The ambiguity characering this category of explanation (ref. Note 1) reduces the reliability that man uses to foresee and control changes to his advantage. In order for LLMs to be reliably applied to a real-world task, it is requisite to integrate LLMs with knowledge databases that contain (1) explicit and interpretable knowledge of isolated facts in the target domain, (2) principles and laws governing interaction between various pieces of information.

As one type of knowledge database, lexical knowledge database supports two scientific explanation categories distinguished in (Hempel, 1965): deductive-nomological explanation and deductive-statistical explanation, both of which involve laws and regularities and interactions among the components inside a language. Lexical knowledge supplies laws and principles governing syntactic-semantic interplay that complements the purely distributional knowledge stored in neural models' parameters and help sensitize the model to more nuanced linguistic patterns (Majewska & Korhonen, 2023). In addition, like other knowledge graphs, lexical knowledge databases are structured, accurate, decisive, interpretable, and dependably domain-specific so that they are highly complementary to LLMs in high-stake scenarios.

**2.2 Verb knowledge database**

Among various type of lexical knowledge database, verb knowledge database attracts the most attention because verbs are pivots in human language and the understanding of the category is crucial in neuroscience, psycholinguistics, artificial intelligence, and other fields (Deng et al., 2023; Kemmerer, 2022; Majewska, 2021). Verb knowledge databases fall into two categories according to the types of information focused in the construction process. One category of verb knowledge databases is built to provide information about events, including places, time, the roles of the participating actors, and the relations among the actors etc. The database built in (Deng et al., 2023) mainly serves this purpose, which specifies semantic features of verbs such as familiarity, agentive information, action effector, perceptual modality, instrumentality, emotional valence, action imageability, action intensity, and usage scenario of action etc. Such dimensions of semantic features can help study the neural representation, neural processing mechanisms, perception modality, and action effector etc. They also help improve the generalization abilities of agents when incorporated into the studies of multi-modal environments. J. Liu et al. (2022) automatically mine verb-oriented commonsense knowledge (e.g., *person eats food*) from large scale corpus with the help of a probabilistic taxonomy.

The other category of verb knowledge databases is built because verbs are clause governors and organizational nuclei in sentence structure (Majewska, 2021). They exhibit semantic and grammatical characteristics that determine the formation of sentence structures. They are pivots where interplay between semantic and syntactic features exhibits regular and meaningful norms that are comprehensible to language users and powerful in predicting linguistic behaviors. This category of knowledge database is built to capture the syntactic-semantic interface information carried by verbs. A lexical resource about verbs can facilitate natural language understanding by mapping verbs to relations over entities expressed by their arguments and adjuncts in the world (Wijaya, 2016).

## 2.3 Towards automatic construction of event knowledge database

A review of the research literature on verb knowledge database shows that there is a trend from focus on minimizing information for lexicon to a full characterization of verb senses so that verb knowledge constitutes knowledge of events that are semi-automatically or automatically acquired.

Traditional studies on verb knowledge construction are guided with the observation that the ideal lexical entry for a word should minimize the information provided for that word, with two characteristics identified in these studies. One characteristic is the focus on arguments and adjuncts for judgment of verb sense. (Levin, 1993, pp. 2-3) focuses native speakers' lexical knowledge of arguments and adjuncts, and describes the knowledge of verbs as the ability to make subtle judgments concerning the occurrence of verbs with a range of arguments and adjuncts in syntactic expressions, the subtle judgments of meaning differences with alternate arguments, and judgments on novel combinations of arguments and adjuncts. Most verb knowledge databases such as FrameNet(Baker, Fillmore, & Lowe, 1998), VerbNet (Schuler, 2005), PropBank (Palmer, Gildea, & Kingsbury, 2005), and Corpus Pattern Analysis (Hanks & Ma, 2020; Hanks & Pustejovsky, 2005) heavily reply on predicate-argument relationships to characterize verb meanings. The semantic frame introduced in Frame Semantics (Charles J Fillmore, 1982) represents verb meanings with a prototypical schema that captures a situation by specifying semantic roles participating in the situation (Charles J. Fillmore, 1976). One example of frame *Abandonment* from the FrameNet database is as follows:

[Another vehicle]_Theme was ABANDONED [at Great Victoria Stree]_Place.

There have also been similar knowledge databases for Chinese such as Chinese FrameNet (郝晓燕, 刘伟, 李茹, & 刘开瑛, 2007) and Chinese VerbNet (M.-C. Liu & Chiang, 2008). One important large-scale verb knowledge database is the Modern Chinese Grammar Information (俞士汶 & 朱学峰, 2017), which contains both morpho-syntactic information of verbs and information of semantic roles, but the morpho-syntactic information is not mapped to semantic roles. Another import verb knowledge database is the Syntactic-Semantic knowledge base of Chinese verbs (袁毓林 & 曹宏, 2022), which maintains that to know how a verb should be used is to know how semantic roles participate to convey semantic meanings and how these semantic roles form patterns in conveying the meanings. One important advantage of this verb knowledge base is the mapping from patterns of semantic roles to semantic meanings.

The other characteristic is the reliance on classification for prediction. As verbs play pivotal roles in language, the construction of verb knowledge database involves one fundamental question in the literature of syntax-semantics interface—**to what extent it is possible to predict syntactic, semantic, or phonological properties of verbs given the knowledge of other verbs.** Levin (1993) argues that the dominant way to address the question is classification, based on the assumption that general meaning components derived of semantically coherent verb classes can be used to predict verbs' syntactic behavior, and proposes to use a range of diathesis alternations such as transitivity, arguments, reflexive pronouns, passive structure, oblique subjects, postverbal subjects, and others to group verbs into semantically coherent classes. Other similar efforts include manner-of-speaking verbs (Zwicky, 1971), change-of-state verbs (Charles J. Fillmore, 1968), and surface-contact verbs (Charles J. Fillmore, 1968). This assumption is accepted and expanded, and is used to design approaches to verb knowledge database construction, such as VerbNet (Schuler, 2005). These approaches differ mainly in two aspects: how to represent these meaning components and how to obtain them. Grouping verbs into finer categories makes it possible to distinguish the subtle differences in both the syntactic behavior and the semantic function of different verbs, and to predict novel verb use. For example, *hit* or *touch* are not change of state verbs, they are not found in causative constructions (such as *The cat touched* and *The door hit*), while *break* is a change of state verb, it occurs in causative constructions (such as *The window broke*) (Levin, 1993).

More recent studies recognize that a full characterization of verb sense goes beyond specification of arguments and adjuncts, and add further requirement for information sufficiency. Generative lexicon argues that a lexical semantics framework should look for representations richer than thematic role description and should include both syntactic structure of the words and the conceptual structures and conceptual domains they operate in (J. Pustejovsky, 1995). Using a decomposition approach, the generative lexicon (James Pustejovsky, 1991; J. Pustejovsky, 1995) proposes four levels of semantic representations: argument structure, event structure, qualia structure, and lexical inheritance structure. Besides argument structure that characterizes thematic roles, the event structure provides information about the internal, subeventual structure, the qualia structure lists the different modes of predication with a lexical item, and lexical inheritance structure identifies how a lexical structure relates to others. The Chinese Verb Library (汪梦翔, 王厚峰, 刘杨, & 饶琪, 2014) propose to include four levels of semantic generalization, i.e., event structure, semantic roles, qualia structure, and syntactic pattern.

Studies in cognitive science propose that actions denoted by verbs aggregates to events and the knowledge of events constitutes part of human cognitive capacity. It is argued that **people organize the explicit knowledge of events, or event schemata, in taxonomies and partonomies** (Zacks & Tversky, 2001). Event schemata are on the one hand discrete, segmented and bounded with beginnings and ends, and on the other hand organized in whole-part hierarchies corresponding to goals and sub-goals of the events (Hard, Tversky, & Lang, 2006). Event schemata drive narrative comprehension (Zacks & Tversky, 2001) by making predictions about what will happen next. (McRae & Matsuki, 2009) and (Metusalem et al., 2012) motivate the concept of **generalized event knowledge**, arguing that theories of sentence comprehension must allow for rapid dynamic interplay between people's knowledge of generalized events such as typical participants, common instruments, time course, and location with the syntactic structure of sentences. (Elman & McRae, 2019) further outline a model of event knowledge with requirements covering thematic roles, hierarchical organization, and **order of events**. First, a piece of event knowledge should provide information supporting inference of activity components not explicitly mentioned or experienced and of higher-order interactions among the activity components. Second, a piece of event knowledge should provide information concerning the temporal structure of multiple activities, or activity sequences, which reflects causal necessity, and order constraints. Third, a piece of event knowledge should indicate typicality, i.e., whether the event is typical or atypical. Events are constructed in a compositional and systematical manner to characterize static or dynamic situations, such as the example of grasping a glass and drink out of it (Butz, 2021).

Recent studies also resort to automatic construction of verb knowledge database. Manual construction of verb knowledge database proves to be time-consuming, costly, and weak in coverage. VerbNet, for instance, suffers from lack of coverage and has no coverage for languages other than English, and expanding coverage through manual effort alone is infeasible (D. W. Peterson & Palmer, 2018). The unsupervised automatic construction of verb knowledge database employs clustering technique to group verbs based on features of shared subcategorization information (Majewska, 2021, pp. 26-27). The subcategorization information is generally obtained by syntactic parsing and the clustering techniques used in the literature include Dirichlet Process Mixture Models (Vlachos, Ghahramani, & Korhonen, 2008; Vlachos, Korhonen, & Ghahramani, 2009), Latent Dirichlet Allocation (Materna, 2012), and Chinese Restaurant Process (Kawahara, Peterson, Popescu, & Palmer, 2014b) etc. For instance, (Kawahara et al., 2014b) use Chinese Restaurant Process to automatically induce verb-specific frames from a massive amount of verb instances. The verb instances are first parsed into dependency trees and the predicate-argument structures are used for the Chinese-Restaurant-Process based clustering. (Materna, 2012) proposes to use Latent Dirichlet Allocation to obtain the probability distribution of semantic roles for each lexical unit using predicate-argument relations such as *subject* and *object*. (D. Peterson, Brown, & Palmer, 2020; D. W. Peterson & Palmer, 2018) take two steps to automatically construct verb knowledge construction: sense induction and verb clustering. The senses of the verbs are first induced by sampling verb use instances with dependency information, and then are clustered together with syntactic patterns to form verb classes. Another approach to verb knowledge database construction is to derive verb clusters from non-expert annotators by exploiting pair-wise similarities, as is exemplified in (Majewska, 2021). The study utilizes spatial arrangement method to collect similarity among verbs and then uses the similarity values to cluster verbs to classes.

# 3 Automatic construction of verb collostruction database

Following the line of automatic construction of verb knowledge database with rich fine-grained information, this paper adopts a new morpho-syntactic form—verb collostruction—as the primary form of verb knowledge and proposes a novel algorithm to automatic generate verb collostructions from large-scale corpora. This section first introduces a formal definition of the concept of verb collostruction, explicates its design features, and explains the generating algorithm based on word embeddings and DBSCAN-based clustering.

## 3.1 Verb collostruction: a formal definition

The concept of verb collostruction is derived from the studies of lexeme-based collostructional analysis Tang (2017); (Tang, 2021; Tang & Liu, 2018), which is a category of collostructional analysis (S. Gries, 2019; Stefan Th. Gries & Stefanowitsch, 2004; Anatol Stefanowitsch & Gries, 2003). A verb collostruction is a construction, a form-meaning pair that presents a particular syntactic pattern of the lexeme (denoted in the Focus slot) in association with a particular communicative function, as is illustrated in Figure 1. It is considered a prototypical and autonomous usage of verb in a language. In line with premise upheld in Construction Grammar that linguistic system is a continuum of successively more abstract constructions from words to idioms to partially filled constructions to abstract constructions (Anatol Stefanowitsch & Gries, 2003), a verb collostruction is a lower-level abstracted construction that is supportive of higher-level generalization such as Frames and verb classes studied in FrameNet and VerbNet. A dynamic inductive process can be used to generalize verb collostructions to frames and verb classes wherein syntactic features or semantic roles are differently weighted and used. In addition, when triggered by linguistic or non-linguistic contexts, people will use the cognitive device of generalization to obtain new frames and verb classes. For instance, the Chinese Verb Library (汪梦翔 et al., 2014) introduces an algorithm that uses Event Structure of verb classification to classify verbs into states, processes, or transitions.

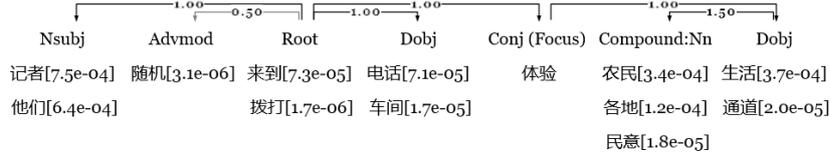

**Figure 1 An illustration of lexeme collostruction**

Assuming dependency grammar (de Marneffe & Nivre, 2019) as the descriptive mechanism of grammar, a lexeme collostruction $C_v$ is a dependency tree embedded with statistical information. Formally, $C_l$ is projective, rooted, ordered, and directed acyclic graph devoted to a target lexeme $l$, as follows:

$$C_l = (S, E, p_{col}), \text{ wherein } S = \{s_i\} \text{ and } E = \{e_j\} \qquad (1)$$

wherein $S$ is a list of ordered slots serving as nodes in the graph. Each $s_i \in S$ contains a set of tuples $\{(w, p_{lex})\}$ wherein $w$ is a collexeme (a lexeme co-occurring with $l$ in $C_l$ in the language) and $p_{lex}$ is the association strength between $w$ and $s_i$. Each $e = (s_i, s_j, r, p_{slot})$ in $E$ is a directed edge from the head slot $s_j$ to the dependent slot $s_i$ in the dependency relation[3] $r$ and $p_{slot}$ is the probability for $e$ to occur in $C_v$. As $C_v$ is projective, the edges in $E$ do not cross each other. Inside $C_v$ there is one special slot $s_v = (l, 1)$ that contains only the target lexeme $l$ and $p_{lex}$ equals 1. This slot is termed the Focus-Slot (shorted as F-Slot).

To make for a lexeme collostruction, Equation (1) requires a further constraint: All the slots inside $S$ should be directly or indirectly linked to the F-Slot so that $S$ falls into two categories, denoted below:

$$S = (S^{child}, S^{ancestor}) \qquad (2)$$

Wherein $S^{child}$ stands for the child slots of F-Slot and the children of the child slots, and $S^{ancestor}$ include the ancestor slot(s) of F-Slot and children of the ancestor slot(s). Such categorization roughly corresponds to independent clauses and dependent clauses distinguished in (Huddleston & Pullum, 2008). $S^{child}$ is mainly used to characterize dependency types inside independent clauses wherein the verbs in question function as predicates, while $S^{ancestor}$ is mainly used to characterize dependency types when the verbs are used in dependent clauses.

---

[3] Dependency relation $r$ as in $r(h, d)$ is used in Dependency Grammar (de Marneffe & Nivre, 2019) to label the syntactic-semantic relation between the head word $h$ and the dependent word $d$. Please refer to Chang, Tseng, Jurafsky, and Manning (2009) and https://github.com/explosion/spacy-models/releases/tag/zh_core_web_trf-3.7.2 for explanation of dependency types used in this study.

When $l$ is a verb, the lexeme collostruction becomes a verb collostruction, as is illustrated in Figure 1 above, wherein $l$ is the Chinese verb 体验'experience denoted by the F-Slot. The slots in $S$ are termed using the edge information. Given an $e = (s_i, s_j, r, p_{slot})$, the dependency type $r$ is used to term $s_j$, the dependent slot, as is illustrated by the dependency relation types CCOMP, DOBJ, ADVMOD, COMPOUND:NN, and DOBJ except for the F-Slot. When a dependency type occurs more than one time, a number is added to distinguish the slot names. The probability of the relation types is marked on the directed edges, and the collexemes and their association strength are also listed for each slot in descending order. The $S^{child}$ of the collostruction includes the slots ADVMOD, DOBJ, and COMPOUND:NN, while the $S^{ancestor}$ includes the slots CCOMP and DOBJ.

On the basis of $S^{child}$ and $S^{ancestor}$, a further classification is made on an edge $e \in E$ according to its role in building or constraining the semantic interpretation of the F-slot, given in Table 2. The No. 1 category, i.e., FOCUS>CHILD, includes all the slots inside $S^{child}$, while No. 2 category includes one slot inside $S^{ancestor}$. From the explanation and illustration in Table 2 it can be observed that No.1-3 categories are more important in determining the semantic function of the F-Slot.

**Table 2 Functional categories of edges according to their relation to F-Slot**

| No. | Category | Explanation | Illustration |
| --- | --- | --- | --- |
| 1 | FOCUS>CHILD | A complement of F-Slot | 帮助**解决**问题。 |
| 2 | HEAD>FOCUS | Immediate context of F-Slot | 组织干部**帮助解决**问题 |
| 3 | CONTEXT>HEAD | Indirect context of F-Slot | **要求**组织干部解决问题 |
| 4 | HEAD>CONTEXT | Relevant context of F-Slot | **组织干部**帮助解决问题 |
| 5 | CONTEXT>CONTEXT | Outskirt context of F-Slot | 组织**全县干部**帮助解决问题 |

### 3.2 Design features of a verb collostruction

As collostructional analysis is specifically geared toward the investigation of lexis-grammar interface (Anatol Stefanowitsch, 2013), a verb collostruction possesses at least the following two design features: functional independence and graded prototypicality with negative evidence, so that it constitutes a piece of event knowledge as is proposed in (Elman & McRae, 2019): information concerning the explicitly and implicitly mentioned components of the action denoted by the verb and the higher-order interactions among these components and information concerning event typicality to support prediction and judgement of novel event combinations. Details follow.

*3.2.1 Functional independence*

*3.2.2 Graded typicality with negative evidence*

For a verb collostruction to support prediction and judgement of a novel event, it is designed to explicitly express graded typicality of events at three levels: collostructional level, slot level, and collexeme level, indicated by the three probability values $p_{col}$, $p_{slot}$ and $p_{lex}$ (ref. Section 2.1). The probability $p_{col}$ indicates how often the verb is used in the particular collostruction the language, $p_{slot}$ indicates the occurrence probability of a particular slot in the collostruction in question, and $p_{lex}$ indicates how strong a collexeme is associated with the collostruction in question.

Graded typicality is also implicitly encoded by the semantic similarity among the collexemes inside each individual slot. In Figure 1, the two collexemes 记者'journalist and 他们'they in the ancestor NSUBJ are semantically similar, while the 生活'life and 通道'passage in the child DOBJ are less similar. In Figure 2, the collexemes inside the ancestor DOBJ and the child DOBJ are also different in mutual semantic similarity. The semantic similarity among the collexemes inside one slot indicates the typicality of how an abstract concept is associated with the slot. From the two example collostructions of 体验'experience, PERSON is a typical concept for the ancestor NSUBJ slot in Figure 1 and the ancestor DOBJ in Figure 2, while LIFE is a less typical concept for the two child DBOJ in the two figures.

This feature of graded typicality, which distinguishes verb collostructions from records of verb knowledge in VerbNet and PropBank, meets several theoretical arguments. Linguistic structures are graded with respect to their degree of prototypicality and cognitive entrenchment (Langacker, 1987, p. 52). Inside a verb collostruction, such graded prototypicality is embodied by typicality at the collocational level, the slot level, and the collexeme level. The including of typicality information in collexeme analysis strongly increases the descriptive adequacy of grammatical description (Anatol Stefanowitsch, 2013; Anatol Stefanowitsch & Gries, 2003), as are observed in various applications of collostruction analysis in corpus linguistics, such as determination of semantic prosody (Anatol Stefanowitsch & Gries, 2003; Tang & Liu, 2018), differentiation of synonymous and alternative constructions (Stefan Th. Gries & Stefanowitsch, 2004), semantic change (Hilpert, 2006, 2007, 2013; Tang & Ye, 2024), language acquisition (Stefan Th Gries & Wulff, 2005, 2009), and acceptability judgment (Anatol Stefanowitsch, 2006, 2008).

Furthermore, the probability information encoded inside a verb collostruction produces prototypicality with negative evidence. (Anatol Stefanowitsch, 2008) and (Anatol Stefanowitsch, 2006) argue that such information does not simply explain the degree of schematicity, conventionality, and grammatical acceptability of existent slots and collexemes inside a verb collostruction, but also provide negative evidence for slots and lexemes that do not occur. Particularly, the collexeme probability, i.e., $p_{lex}$ is computed on the basis of contingency table that include co-occurrence information of collostructs (syntactic patterns of the collostructions) and collexemes (S. Gries, 2019). The statistical values yielded from this computation are evidence for null hypothesis—hypothesis that two features do not co-occur. The higher these values, the stronger the expectation of co-occurrence, and the stronger of negative evidence (Anatol Stefanowitsch, 2008). Strong expectation of co-occurrences between collexemes and collostructions constitutes strong preemption that can be used for judgment of acceptability and grammaticality. Given the statistical principle of Law of Large Numbers (Ross, Ross, Ross, & Ross, 1976), the use of large corpus and the exhaustive extraction of all occurrences of the target grammatical phenomenon enhances the validity of negative evidence (Anatol Stefanowitsch, 2006). The collostructions obtained this way should be closer to the true usage distribution of the verbs in the language.

Armed with this feature of graded typicality, verb collostructions provide information <span style="color:red">needed in explaining the process of sentence comprehension, the probabilistic information among the arguments in particular</span>. In sentence comprehension models (Gibson & Pearlmutter, 1998; MacDonald, Pearlmutter, & Seidenberg, 1994), lexical processing involves activating different types of information associated with a wordform including phonological form, orthographic form, semantics, grammatical features, and the grammatical and probabilistic relationships such as locality (distance among two constituents) and phrasal-level contingent frequency that hold among these features. The above-mentioned probability values provide exactly the information specified in these comprehension models.

## 3.3 Algorithm of verb collostruction generation

This section explains the algorithm that generates verb collostructions from large-scale corpus for a given verb, given in Figure 3. After obtaining for the verb sentence instances from a monolingual corpus, it takes three steps of clustering to generate collostructions from input sentence instances. In the first step, instance sentences are converted to sentence embeddings with sentence-BERT (Reimers & Gurevych, 2019) and are clustered to obtain semantically similar sentence groups. For each obtained sentence cluster, the second clustering step parses the sentences inside the cluster, retrieve clause structures from the sentences, and cluster the clause structures based on syntactic and semantic similarity among them. In the third step, the obtained clusters are then used to generate collostructions except for outliers. Rationales are detailed below.

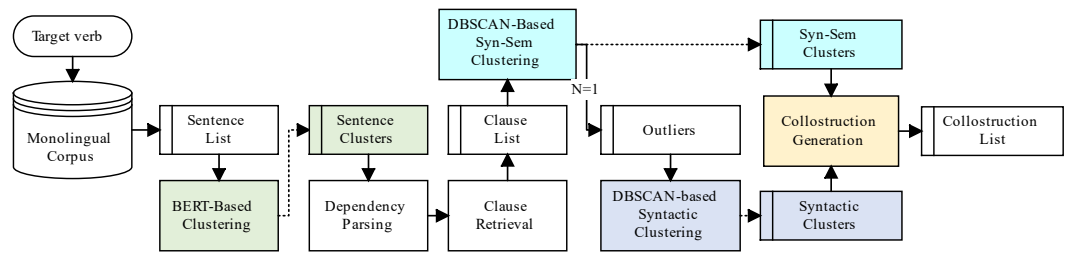

**Figure 3 Automatic generation of verb collostructions**

### 3.3.1 BERT-based clustering for verb sense distinction

As verbs are often polysemous, sense distinction is required for collostruction generation so as to meet the design feature of functional independence. This is achieved by BERT-based clustering in the present study. The sentences are first converted to embeddings using a pretrained BERT-based model[4] so that semantic similarity can be computed between sentences and clustering algorithms can be applied to group input sentences into different clusters. Each cluster stands for one particular context in which the target verb is used, and the target verb used within one particular context is expected to share one sense. Therefore, when the clause structures associated with the target verb are used to generate collostructions, the collexemes inside one collostruction should be semantically compatible with the sense of the target verb and the collostruction in question should meet the feature of functional independence.

---
[4] BGE-M3 (Chen et al., 2024) is used in the present study. AgglomerativeClustering from sk-learn is used for clustering.

*3.3.2 Clause Structure Retrieval*

Verb collostructions are not directly generated from the dependency tree of a full sentence, but from a clause structure retrieved from the tree. This strategy is adopted for two rationales. The first is the assumption that a verb sense can be deterministic with information gathered from within a clause (ref. Section 2.2.1). In clause structure retrieval, it is required that sufficient information be gathered from an input sentence so that the sense judgment of the verb is deterministic.

More specifically, we formally define a clause with Equation (3) below:

$$Clause = (V^{child}, V^{ancestor}, v_t) \quad (3)$$

wherein $V^{child}$ stands for the indexed child nodes of the target verb node $v_t$ inside a dependency tree and $V^{ancestor}$ includes the indexed ancestor node of $v_t$ and skeleton nodes associated with the ancestor node (explained below). Each $v$ in $V^{child}$ or $V^{ancestor}$ is of the form $r(w_h, w_d)$ wherein $r$ stands for dependency relation type, $w_h$ is the head word and $w_d$ is the dependent word. As such, a clause structure is a subgraph of the dependency tree from which it is obtained. To ensure that the retrieved clause is semantically well-formed, this study uses the following strategies to retrieve $V^{child}$ and $V^{ancestor}$:

(1) If the target verb is the main predicate in a sentence, $V_{child}$ includes all the child nodes of the target verb in the dependency tree except for the conj, and $V_{ancestor}$ includes the immediate ancestor of the target verb (for independent clause);

(2) Otherwise, if the verb is in conjunction with other verbs and its children nodes contains a subject, $V_{child}$ includes all the child nodes of the target verb in the dependency tree, and $V_{ancestor}$ includes the immediate ancestor of the target verb (for dependent clause type 1);

(3) Otherwise, if the ancestor of the target verb (***ancestor-01***) contains a subject, $V_{child}$ includes all the child nodes of the ancestor in the dependency tree, and $V_{ancestor}$ includes the immediate ancestor of ***ancestor-01*** (for dependent clause type 2);

(4) Otherwise, obtain the ancestor of ***ancestor-01*** as ***ancestor-02***, $V_{child}$ includes all the child nodes of ***ancestor-02*** in the dependency tree and $V_{ancestor}$ includes the immediate ancestor of ***ancestor-02*** (for dependent clause type 3).

### 3.3.3 DBSCAN-based clustering

Both DBSCAN-based Syn-Sem Clustering and DBSCAN-based Syntactic Clustering use the clustering algorithm proposed in DepCluster (Tang, 2017) to cluster the clause structures. The proposed algorithm in Tang (2017) makes two modifications to the original DBSCAN algorithm (Ester, Kriegel, Sander, & Xu, 1996). The first modification concerns epsilon, the threshold used to decide whether one clause structure is a neighbor of another. Instead of using the same value for all the clause structures, DepCluster computes the epsilon for each clause structure utilizing the minimal distance between the clause structure and other clause structures participating in the clustering process. The second modification concerns how neighborhood is formed. A clause structure is not allowed to enter the neighborhood unless its distance to each clause structure already inside the neighborhood is smaller to the epsilon of that clause structure.

Nevertheless, DBSCAN-based Syn-Sem Clustering and DBSCAN-based Syntactic Clustering differ in how the distance between two clause structures are computed. Given two clause structures $C_1$ and $C_1$, the distance between them $d(C_1, C_2)$ is computed via the similarity between them $s(C_1, C_2)$ with the assumption that 0.05 is the minimum similarity between two clause structures, as follows:

$$d(C_1, C_2) = |\frac{\ln(sim(C_1, C_2))}{\ln(0.05)}| \qquad (4)$$

$$sim(C_1, C_2) = \alpha \times sim(V_1^{child}, V_2^{child}) + \beta \times sim(V_1^{ancestor}, V_2^{ancestor}) \qquad (5)$$

As indicated in Equation (5), the similarity computation between two clause structures uses two constants $\alpha$ and $\beta$ to regulate the weights of child nodes and ancestor nodes inside the clause structures. As both $V^{child}$ and $V^{ancestor}$ contain indexed nodes of the form $r(w_h, w_d)$, $sim(V_1, V_2)$ can be computed using dynamic programming on the basis of the similarity between two nodes $sim(n^1, n^2)$, as follows:

$$sim(n^1, n^2) = sim(r^1(w_h^1, w_d^1), r^2(w_h^2, w_d^2)) \qquad (6)$$

That is, the similarity between two nodes depends on whether they share the same dependency relation $r^1 = r^2$, on the similarity between the head words $w_h^1$ and $w_h^2$, and on the similarity between the dependent words $w_d^1$ and $w_d^2$. In DBSCAN-based Syn-Sem Clustering, all the three criteria are used to compute the between-node similarity, while in DBSCAN-based Syntactic Clustering, only the criterion of dependency identity is used to compute the between-node similarity.

*3.3.4 Collostruction generation*

The clause structure clusters yielded in both DBSCAN-based Syn-Sem Clustering and DBSCAN-based Syntactic Clustering are used to generate collostructions. One cluster of clause structures generates one collostruction. As the clause structures within one cluster share similar syntactic structures (and semantic function with DBSCAN-based Syn-Sem Clustering), the number of clause structures is an indicator of the degree of prototypicality of the generated collostruction in comparison with other clusters obtained in the clustering process.

Because both a clause structure and a collostruction are graphs (ref. Section 2.1), collostruction generation is essentially a process of merging clause structures inside a cluster into one graph and finding the most representative ordered, projective, rooted, and directed acyclic subgraph inside the merged graph. This procedure consists of three steps: generating a directed graph using linear adjacency, finding traversal paths with starting nodes in the clause structures, and selecting the best traversal path with a set of constraints.

The first step is to generate from the cluster of clause structures a new directed graph using linear adjacency as the edge instead of the dependency relationship. As a clause structure is an ordered graph, there is linear order among the nodes (or words, denoted by dependency relation that it forms with its head) inside the graph. Using the nodes in the clause structures and linear adjacency between nodes as edges, a new directed graph $G_{order}$ can be generated from the cluster of clause structures, wherein the weight of an edge is the frequency of the adjacency pair in the clause structure clusters.

The second step is to obtain a list of traversal paths $L_{order}$ from $G_{order}$ using all the starting nodes of the clause structures in the cluster. For each starting node, a depth-first search is used to obtain a traversal path inside $G_{order}$. Therefore, the obtained list of traversal paths stands for all possible linear arrangement of the nodes inside $G_{order}$.

The third step is to select from $L_{order}$ the best path using a set of constraints. First, the following two constraints are used to qualify a traversal path: (1) Use the node of the target verb as the anchor and check the left side and right side of the target verb. If the traversal path contains a node that is expected to occur to the left side but is found in the right side or vice versa, the traversal path is removed from $L_{order}$; (2) If the traversal path does not contain the node of the target verb and the ancestor node of the target verb node. Second, rank the traversal paths inside $L_{order}$ according to the priority score computed with the following equation:

$$Score = \frac{Coverage + Average}{1 + NumDangle} \tag{7}$$

wherein $NumDangle$ is the number of dangling nodes in the path—the nodes that form no dependency relationship with any other nodes in the path, $Coverage$ is the ratio of the number of non-dangling nodes against the length of the path, and $Average$ is the average edge weight among the non-dangling nodes in $G_{order}$.

The selected traversal path is then used to generate a collostruction, together with frequency information gathered from the clause structure cluster and the corpus. Following the Bayesian Theorem, the association strength between a collexeme and a collostruction $\rho_{collexeme}$ is computed as the posterior probability, i.e.,

$$\rho_{collexeme} = p(collexeme|collostr) = \frac{p(collexem, collostr) \times p(collexeme)}{p(collostr)} \tag{8}$$

Wherein $p(\cdot)$ stands for probability or conditional probability.

### 3.4 Experiment configuration

*3.4.1 Corpus*

The monolingual corpus of Chinese used for collostruction retrieval contains 474,432,680 tokens derived from two sub-corpora. The first sub-corpus contains 1956-2012 texts from the newspaper *People's Daily*. The second sub-corpus is the *Modern Chinese Corpus* compiled by the National Language Committee of China. We argue that the combined corpus is balanced in genres because the newspaper *People's Daily* publishes news of multiple disciplines and industries, comments, and stories, and the *Modern Chinese Corpus* itself is also a balanced corpus.

*3.4.2 Implementation details*

The algorithm in Figure is implemented with Python scripts, making use of four Python libraries: *sentence transformers*, *sk-learn, spaCy*, and *transformers*[5]. The library of *sentence transformers* is used in Section 3.1 to obtain sentence embeddings and the clustering algorithm is *AgglomerativeClustering* from *sk-learn* with the threshold cosine similarity 0.5. The library *spaCy* is mainly used for dependency parsing. The library *transformers* is used to obtain word embeddings for Chinese words to support similarity computation between words in Equation (5).

## 4 Statistical analysis of verb collostruction database

Inside the collostruction generating algorithm in Figure 1, the two clustering components, namely the Bert-Based Clustering component and the DBSCAN-Based Syn-Sem Clustering component play the vital role in generating collostructions from input sentences for each verb. The Bert-Based Clustering component ensures that the sentences inside a cluster fall into one sub-language domain with mutual similarity among the instances larger than 0.5. The DBSCAN-Based Syn-Sem Clustering component further restricts the clauses inside a cluster to be mutually similar in both syntax and semantics. Accordingly, the collostructions generated with the above two clustering processes for a given verb exhibit two characteristics: (1) The collostructions of a verb form a fractal; (2) Each collostruction is functionally independent.

### 4.1 From typicality to fractal

Several studies affirm that languages should present a fractal structure with the property of cascading self-similarity, i.e., the structures of linguistic objects are scale invariant (e.g., Mandelbrot (1977), (Ribeiro, Bernardes, & Mello, 2023), Hrebíček (1994), (Andres, 2010), and (Tang & Ye, 2024)). Statistical data obtained in the experiments also yield evidence supporting the above observation, that is, a verb is a fractal with cascading self-similarity. The usage frequency of senses of a verb is in power law distribution. Simultaneously, the usage frequency of collostructions of each sense of the verb is also in power law distribution. The power-law distribution cascades from verb sense to collostruction, forming a cascading similarity.

---

[5] The Python scripts are to be released on github (https://github.com/). The versions and pretrained models of dependent python libraries are as follows: spaCy, version 3.7.0, with the pretrained model "zh_core_web_trf-3.7.2"; sentence-transformers, version 2.2.2, with pretrained model "BGE-M3"; sk-learn, version 1.4.0; transformers, version 4.41.2, with pretrained model chinese-roberta-wwm-ext.

Take the verb 上升'rise for illustration. Applying the Bert-Based Clustering component to 40,000 sample sentences of the verb obtains 767 clusters, among which 147 clusters generate at least one collostruction. Figure 4 gives the percentages of the 147 clusters in the 40000 samples obtained by sum the percentages of the collostructions of each cluster[6], together with power law regression[7]. The R value (the likelihood ratio for the data to be in power law regression), the p value (chance result of data fluctuation), and visual judgment of the curve in the figure show that the percentages of the sense clusters are in a power law distribution.

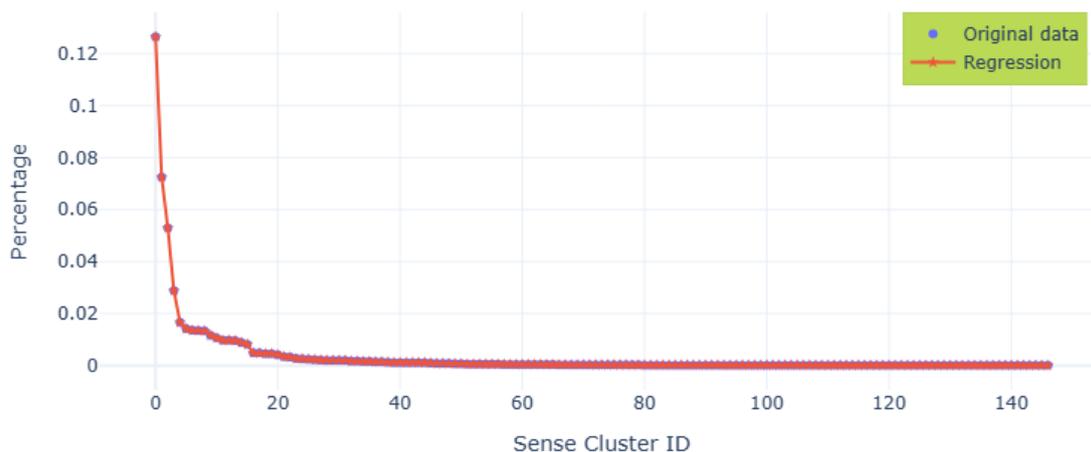

**Figure 4 Percentages of sense clusters in power law distribution. R=5.518, and p=3.426E-08**

Take the most dominant sense in Figure 4 for further analysis. The statistical distribution of the percentages of the collostructions of the sense is give in Figure 5, exhibiting the characteristics of a power law distribution. With the assumption that a verb consists of senses and a sense of a verb consists of collostructions, the shared power law distribution among the senses of the verb and the collostructions of one sense evidence cascading self-similarity.

---

[6] Those clusters that failed to generate any collostruction are not included in the current discussion as these clusters are statistically insignificant. Failure to generate any collostruction within a cluster implies that no similar syntactic and semantic pattern is found among the instance sentences in the cluster, very often due to a insignificant frequency of the cluster.

[7] The power law regression is obtained with the python package powerlaw (Alstott, Bullmore, & Plenz, 2014). Applying the function distribution_compare('power_law', 'exponential') return two parameters: R and p. R is the likelihood ratio between the distribution of power law and exponential. It will be positive if the data is more likely in power law distribution. A higher R indicates higher likelihood for the data to be in power law distribution. The parameter p is the significance value indicates the probability of data fluctuation. A lower p indicates less chance of data fluctuation and more certainty.

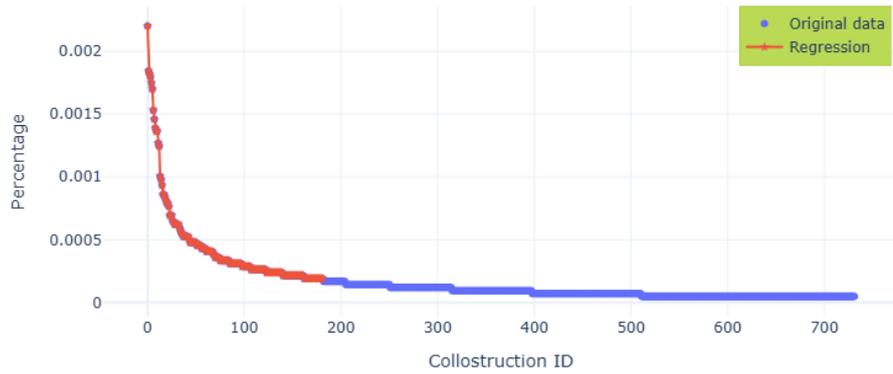

**Figure 5 Collostruction percentages in power law distribution. R=2.373, and p=7.181E-05.**

The same analytical method is applied to all the rest 97 verbs, yielding the sense-level R values, the collostruction-level R values, the sense-level p values and the collostruction-level p values given in Figure 6. The average sense-level R is about 2, the collostruction-level R values is about 3, while the average p values are all very small. The frequency of senses of these verbs are in the power law distribution, and the frequency of collostructions of each sense of the verbs are also in the power law distribution. The statistical structure of the collostructions of one verb sense is similar to the statistical structure of the senses of the verb. These data prove that generally the structure of a verb is statistically self-similar in terms of senses and morpho-syntactic patterns expressing these senses. A verb is a fractal.

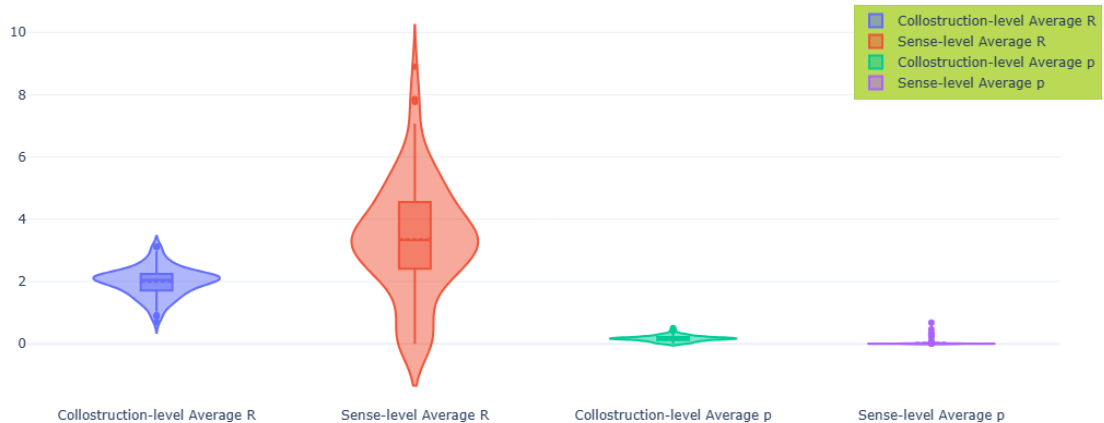

**Figure 6 R values of collostruction-level percentages and sense-level percentages of all verbs**

## 4.2 Functional independence

### 4.2.1 Explicit semantic components

Explicit semantic components provide important information about thematic roles,

subcategorization, and action manners to facilitate sense judgment. Figure 7 plots the average occurrence probability and the average $p_{slot}$ (denoted by Slot P) of the majority dependency types that denote explicit semantic components collected from the collostructions of all the 98 verbs. The top three child slots with the highest occurrence probability are DOBJ, DNSUBJ, and DADVMOD, supporting that observation that participating entities and adjuncts play important roles in characterizing verbs. Nevertheless, statistics in the figure also show that such information is not always present inside a sentence. The dependency type DOBJ occurs in about 50% of all the collostructions, while NSUBJ occurs in about 40% of the collostructions. The ancestor slot ACL occurs in less than 10% of the collostructions. Furthermore, note that some auxiliary dependency types such as AUX:ASP, AUX:MODAL occur relatively frequent as compared with adjuncts, indicating the information of aspects and modals is closely associated with verb usage.

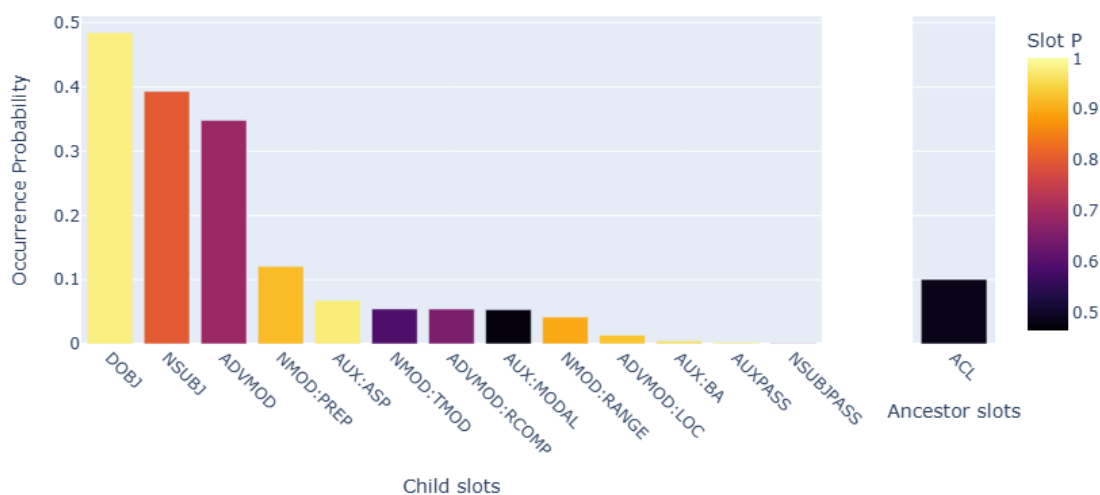

**Figure 7 Percentages and Weights of sense-specifying dependency types**

The statistics of $p_{slot}$ indicate the probability of a dependency type inside a collostruction when it occurs in it. The average $p_{slot}$ values of AUX:ASP, AUX:BA, and AUXPASS that encodes syntactic information are higher than 0.9, indicating strong presence of these dependency types in some collostructions and strong interactions of the semantic components inside the collostructions, as is illustrated in Figure 8 below. In the figure the ADVMOD slot, in which the collexemes 曾经'ever and 从不'never denotes the past time, interacts with AUX:ASP to express a past tense. Furthermore, the higher $p_{slot}$ value of DOBJ than NSUBJ shows that DOBJ is more likely an indicator of verb sense than NSUBJ.

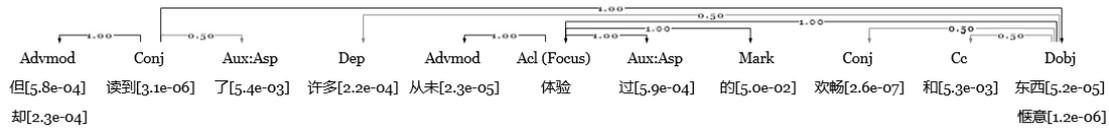

**Figure 8 Illustration of high average $p_{slot}$ dependency types**

### 4.2.2 Prototypical Action Sequence

Using mappings between slots and action sequence and the information of typicality obtained for individual verbs, prototypical action sequences can be generated for each verb, which support further inquiry into cognitive regularities associated with the particular language.

Within a collostruction list of a verb, the process of obtaining the prototypical action sequences is as follows:

(1) Obtain all the slots of the type $s_i$ with the following specification:

$e = (s_i, \text{F-Slot}, r, p_{slot})$ and $r \in \{\text{XCOMP}, \text{NSUBJ}, \text{DOBJ}, \text{COMPOUND: VC}, \ldots\}$

or $\quad e = (\text{F-Slot}, s_j, r, p_{slot})$ and $r \in \{\text{XCOMP}, \text{CONJ}\}$

(2) Obtain all possible sememes associated with each collexeme and hypernyms of each obtained sememes

(3) Sort the sememes according to their frequency and retrieve the five sememes with the highest frequency as prototypical action sequence.

Table 4 gives the list of the prototypical action sequences obtained for the verb 结婚'marry. From the action sequences it can be inferred that in Chinese language the action of marriage is strongly characterized with subjectivity (observed in sememes like willing|愿意, fit|适合, plan|计划, and persuade|劝说) and administration (observed in sememes like prohibit|禁止, record|记录, manage|管理). Such characteristics can be used in further studies in fields like language and culture and cross-culture comparison.

**Table 4 List the prototypical action sequences for 结婚'marry**

| Dependency name | Action sequences | Sentence Example |
|---|---|---|
| CHILD: XCOMP | ('willing|愿意', 18), ('fit|适合', 3) | 不敢结婚、更不敢要孩子。 |
| ANCESTOR: CCOMP | ('plan|计划', 52), ('willing|愿意', 37), ('exist|存在', 20), ('persuade|劝说', 18), ('prohibit|禁止', 13) | 一些年青人将准备结婚的钱也拿了出来。 |
| ANCESTOR: OMPOUND:VC | ('record|记录', 9) | 他俩登记结婚了。 |
| ANCESTOR: NMOD:PREP | ('undergo|经受', 11), ('manage|管理', 3), ('arrive|到达', 3), | 有 3.3％的妇女因为结婚而失 |

|  |  |  |
|---|---|---|
|  | ('ResultIn|导致', 3), ('buy|买', 3) | 去土地。 |
| **ANCESTOR: DOBJ** | ('evade|回避', 3), ('prohibit|禁止', 3) | 一是避免近亲结婚。 |
| **ANCESTOR: NSUBJ** | ('be|是', 3) |  |

*4.2.3 Within slot collexeme similarity*

The information compatibility of a collostruction is mainly observed in the semantic coherence within the collexemes inside each slot, measured by average semantic similarity among the collexemes. Table 5 below illustrates the average semantic similarity of the collexemes of the present slots in Figure 1, computed with the formula $\overline{sim} = \frac{\sum_{i,j} cosine(e_i, e_j)}{N-1}$, wherein $e_i$ and $e_j$ are word beddings of two collexemes of the slot in question and $N$ is the number of collexemes in the slot. Figure X gives the average similarity of all slot types out of all the collostructions of the 100 verbs. It can be observed that all the slots have similarity values bigger than 0.8, indicating that the collexemes inside a slot generally form a coherent concept so that the collostruction is functionally independent.

**Table 5 Illustration of average within-slot semantic similarity among the collexemes**

| Slot Name | Collexemes | Average Similarity |
|---|---|---|
| Ccomp | 出'exit, 发现'discover, 前往'approach, 欣赏'admire | 0.853 |
| Advmod | 深度'in-depth, 即可'immediately | 0.886 |
| Dobj | 中国'China, 文化'culture, 生活'life, 新加坡'Singapore, 风情'charms | 0.842 |

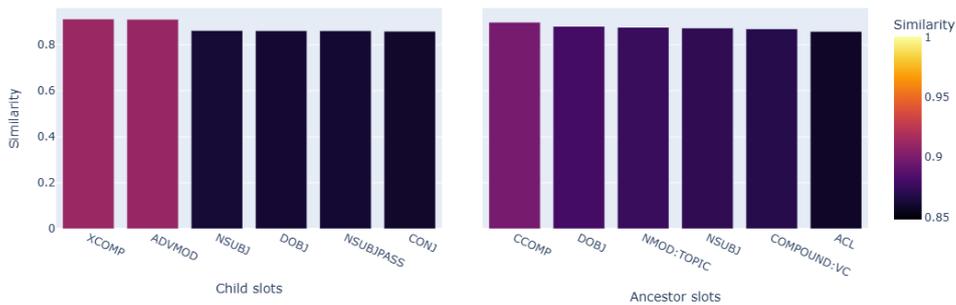

**Figure 9 Average within-slot similarity in verb collostruction database**

# 5 Evaluation with grammatical error correction

Grammatical error correction (Bryant et al., 2023), the error correction of verbs in particular, can be used to test the design feature of graded prototypicality of verb collostructions. A verb collostruction is a prototypical patten that encodes a particular semantic function with explicit syntactic and lexical information, and a conventional pattern that carries negative evidence for unacceptable grammatical forms. From a language learning perspective, verb collostructions are accepted templates in a language community and can be used for two purposes: acceptability check and usage reference. Both purposes are achieved via similarity computation. An expression can be checked against these templates and sharp disparity from verb collostructions symbolizes grammatical error(s). The collostructions with the maximum similarity to the input expression can be the on-the-point references for language learners, prototypicality of which enables the interpretability of the error-check results for language learners as is advocated in Kaneko, Takase, Niwa, and Okazaki (2022).

This section proposes a supervised approach that uses verb collostructions for error-detection. Nevertheless, the proposed approach differs from the category of statistical classifier specified in Bryant et al. (2023), it judges the usage of a verb inside an input sentence by comparing the usage with the collostructions of the verb inside the collostruction database, as is illustrated in Figure 9. A verb grammar error dataset is used to train a neural model of verb error detection that yields a tuple (C-prob, E-Prob), with C-prob for correct-probability and E-Prob for error-probability. A usage is considered an error if $\text{C-prob} - \text{E-Prob} < 0$. The input to the neural model is a feature vector obtained by searching for collostructions inside the collostruction dataset that are the most similar to the clause structure containing the verb retrieved from the input sentence. As the procedure of clause retrieval is already explained in Section 3.2, the following sections are focused on the procedure of searching for max similarity and the construction of the neural detection model.

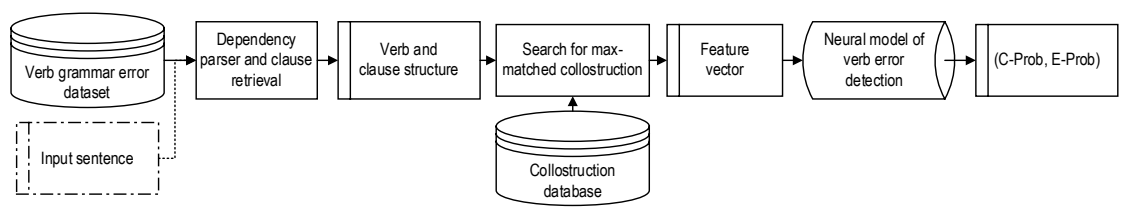

**Figure 9 Verb error detection framework for grammatical error correction. C-prob stands for Correct Probability and E-Prob for Error Probability.**

## 5.1 Searching for maximum-matched collostruction

The component of similarity computation takes two inputs—the clause structure retrieved from a target sentence and the target verb identified in the sentence, use the inputs to search the collostruction database for best-match collostructions, and generate a feature vector as the input to the neural error detection model.

The searching procedure consists of two steps. The first step is to search for best match collostructions with heuristic patterns for a maximum of 12 collostructions of a target verb that best match the semantic function encoded inside the retrieved clause. With the assumptions that the semantic and syntactic functions of a clause can be decomposed into continuity of bigrams, adjacent dependency pairs, and word-dependency pairs, four heuristic patterns are used to search for collocations including bi-word-dependencies, bigrams and unigrams, bi-dependencies, and uni-word-dependency retrieved from the input clause structure, as are given in Table 6. For each pattern, three collostructions that have the maximum matches are collected.

**Table 6 Heuristic searching patterns**

| No. | Pattern Category | Structure | Illustration |
|---|---|---|---|
| 1 | bi-word-dependency set | $\{[(word_i, dep_i), (word_{i+1}, dep_{i+1})]_k\}$ | [(*beautiful*, amod), (*flower*, dobj), …] |
| 2 | bi-gram and unigram set | $\{(word_i, word_{i+1})_k\}, \{word_i\}$ | [(beautiful, flower), …, beautiful, flower, …] |
| 4 | bi-dependency set | $\{(dep_i, dep_{i+1})_k\}$ | [(amod, dobj), …] |
| 3 | uni-word-dependency set | $\{(word_i, dep_i)\}$ | [(beautiful, amod), …] |

The second searching step is to select the collocation that maximumly match the input clause from the collostructions obtained by the above heuristic search. Given a *clause* and a collostruction $C$, the match between them is computed with Equation (9) below:

$$C_{top} = argmax_C^{aSim_{2clause}+bSim_{2col}+cCov_{clause}+dDen_{clause}+eDen_{col}}, \; a+b+c+d+e = 1 \quad (9)$$

wherein $Sim_{2clause}$, $Sim_{2col}$ are the asymmetric similarity metrics computed between *clause* and $C$, $Cov_{clause}$ is the percentage of matched dependencies in *clause*, and $Den_{clause}$ and $Den_{col}$ are respectively the continuity of matched dependencies in *clause* and $C$, which are detailed below.

To compute the asymmetric similarities between $Sim_{2clause}$ and $Sim_{2col}$, the dynamic programming is used to align the slots in *clause* and the slots in $C$, based on slot similarity. The similarity between one slot $r^{cls}(w_h^{cls}, w_d^{cls})$ in *clause* and one slot $(r^{col}(W_h^{col}, W_d^{col}), p)$ in $C$ is based on a fuzzy-match because it is likely for *clause* to contain grammatical errors and to be improperly parsed by a dependency parser, leading to improper dependency triples. The fuzzy match

is explained below:

$$sim(r^{cls}, r^{col}) = \text{minimum-edit-distance}(r^{cls}, r^{col}) \tag{10}$$

$$sim(w_h^{cls}, W_h^{col}) = argmax_{w_h^{col} \in W_h^{col}} sim(w_h^{cls}, w_h^{col}) \tag{11}$$

$$sim(w_d^{cls}, W_d^{col}) = argmax_{w_d^{col} \in W_d^{col}} sim(w_d^{cls}, w_d^{col}) \tag{12}$$

$$sim(cls, colloc) = p \times sim(r^{cls}, r^{col}) \times (\alpha \times sim(w_h^{cls}, W_h^{col}) + \beta \times sim(w_d^{cls}, W_d^{col})) \tag{13}$$

The dynamic programming applied to the clause structure and a collostruction will yield the best alignment $A$ between *clause* and $C$, with a list of similarity values $[a_1 = sim(cls_i, col_j), a_2 = sim(cls_l, col_k), \dots]$ that matches a list of slot indices in the clause structure and a different list of slot indices in the collostruction. Following Tversky similarity (Tversky, 1977), the asymmetric similarity between *clause* and $C$ is computed based on the similarity list. The $Sim_{2col}$, i.e., the similarity against $C$ that measures how similar the clause structure is to the collostruction, is computed with Equation (12) below:

$$Z = \sum_{a_i \in A} a_i \tag{14}$$

$$Sim_{2col} = \frac{Z}{Z + 0.1 \times (M-Z) + 0.9 \times (N-Z)} \tag{15}$$

wherein $M$ is the number of slots in the *clause* and $N$ is the number of slots in $C$. The $Sim_{2clause}$ is the similarity against *clause* that measures how similar $C$ is to *clause*, computed with different constant weights as Equation (16), given below:

$$Sim_{2clause} = \frac{Z}{Z + 0.9 \times (M-Z) + 0.1 \times (N-Z)} \tag{16}$$

Both the match coverage and continuity in *clause* and $C$ are computed with the best alignment $A$, as follows:

$$Cov_{clause} = \frac{Length\ of\ A}{Length\ of\ clause} \tag{17}$$

$$Den_{clause} = \frac{Number\ of\ adjacent\ clause\ slot\ (slot_i, slot_{i+1})\ in\ A}{Length\ of\ clause} \tag{18}$$

$$Den_{clause} = \frac{Number\ of\ adjacent\ collostruction\ slot\ (slot_i, slot_{i+1})\ in\ A}{Length\ of\ C} \tag{19}$$

### 5.2 Feature Vector

From the $C_{top}$, *clause*, and $A_{top}$ obtained above, the features as are specified in Table 10 are generated, wherein the Example column is from the *clause* retrieved from Example 1 and the

$C_{top}$ in Figure 10. Note that for both $C_{top}$ and *clause*, only the FOCUS>CHILD and HEAD>FOCUS dependency types, i.e. constituents that are immediately attached to the Focus verb are used as input features. The similarity values in the brackets are obtained from $A_{top}$. As the collostructions are generated from clusters of clause structures in language use, the idea behind the generated feature vector is to measure how *clause* resembles $C_{top}$.

**Table 10 Feature vector with illustrations**

| No. | Feature name | Explanation | Example |
|---|---|---|---|
| 1 | CORE-DEP-in-COL | The dependency type of the key verb in $C_{top}$; | Dep |
| 2 | DEPs-in-COL | List of FOCUS>CHILD and HEAD>FOCUS dependency types in $C_{top}$ with alignment information | [Acl(推动等,0.0), Nsubj(成本等,1.0), Ccomp(上升等,0.39), Mark(的,0.0), Root(原因等,0.28)] |
| 3 | CORE-DEP-in-CLS | The dependency type of the key verb in *clause*; | Dep |
| 4 | DEPs-in-CLS | List of FOCUS>CHILD and HEAD>FOCUS dependency types in *clause* with alignment information | [Conj(抱,0), Punct(,,0), Dep(导致,0), Nmod:tmod(今日,0.374), Nsubj(离婚率,0.589), Ccomp(上升,0.009), advmod(都,0.0), Root(有,0), Dobj(损失,0)] |

Example 1 待成人后抱着孤芳自赏、持傲不羁的态度，两性的偏见扩大，导致今日离婚率上升，都有莫大的损失。

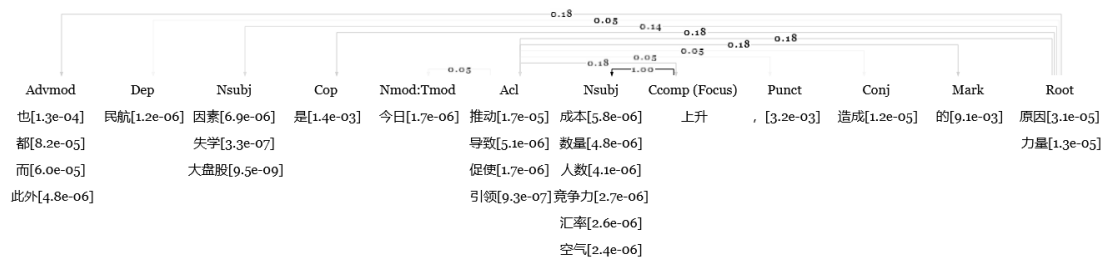

**Figure 10 An illustration of $C_{top}$ with 上升'rise as focus verb**

### 5.3 Neural model of verb error detection

Figure 11 gives the neural network model that takes the features in Table 10 as input for verb error detection. The general idea behind the model is to use Multi-Head-Attention to transform the dependency types in CORE-DEP-in-COL, CORE-DEP-in-CLS, DEPs-in-COL, and DEPs-in-CLS in combination with similarity values CORE-DEP-in-CLS, DEPs-in-COL to a pattern to support CNN-based pattern recognition. The model consists of two components: Feature Transformation and Pattern Recognition. In the component of Feature Transformation, bidirectional long-short time

Memory (LSTM), transformer encoder, and MultiHeadAttention are used to respectively transform features of CORE-DEP-in-COL and DEPs-in-COL, and features of CORE-DEP-in-CLS and DEPs-in-CLS after embedding conversion. The obtained tensors are then respectively integrated with the input similarity feature vectors contained in DEPs-in-COL and DEPs-in-CLS using MultiHeadAttention. Afterwards, the two obtained tensors are further integrated using cross attention with MultiHeadAttention to form an input tensor to the component of pattern recognition.

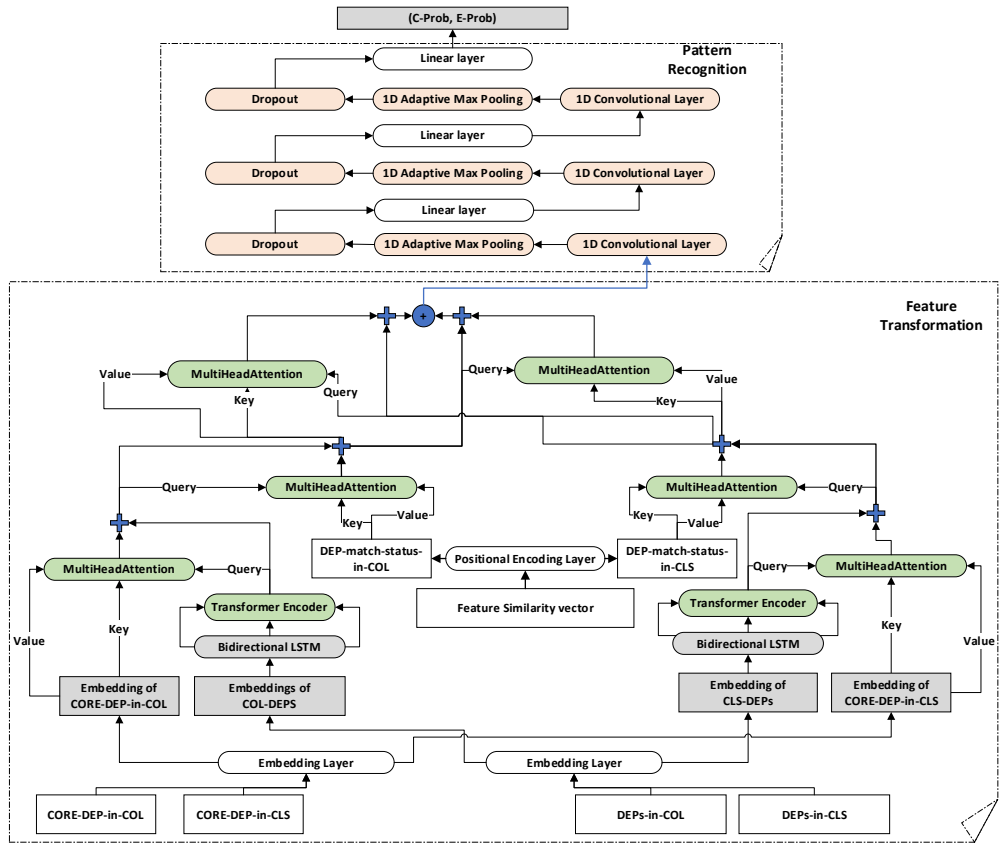

**Figure 11 Neural Network Based Grammatical Error Identification Model. The + symbol with circle means concatenation and the bigger + means tensor addition.**

The component of Pattern Recognition consists of three convolutional layers. It takes the outcome from Unmatched Feature Transformation and outputs the final outcome (C-Prob, E-Prob). If $C\text{-Prob} - E\text{-Prob} > 0$, the clause structure is considered grammatically correct, otherwise it is grammatically ill-formed.

### 5.4 Experiment configuration

For evaluation purpose, 100 Chinese verbs are selected from the 2016-2018 training data for

Chinese Grammatical Error Diagnosis (CGED) (Rao, Gong, Zhang, & Xun, 2018) by first ranking the verbs in the CGED data according to their frequency and then by random sampling. The selected verbs include both high frequency verbs (with 3535 as the maximum) and low frequency verbs (with 4 as the minimum).

The data for grammatical error correction are also retrieved from the above CGED data using the following procedure: (1) From the data retrieve sentences that contain the selected verbs together with their associated annotations; (2) Convert each sentence with their annotation to the form illustrated in Example (2) below:

Example 2

Original text：上女子中学的人，大学毕业后会发生不知道怎么男性交往。

Correction：上女子中学的人，大学毕业后会不知道怎么跟男性交往。

Errors: (1) begin-offset: 23; end-offset: 25; error-type: error

The information in the Errors specifies the indexed position of the verb 交往'interact in the original text and the error type, which is error, denoting a grammatical mistake. In order to prevent high frequency verbs from dominating the obtained data, only 200 instances are sampled for those verbs with frequency higher than 200. The final data contain 3863 instances and are split to training data, test data, and evaluation data according to the ratio 70:15:15. Also note that the data is highly unbalanced with 32.4% correct usage and 77.6% incorrect usage.

The algorithm in Figure 2 is also implemented with Python scripts, using the same Python libraries introduced in Section 4.2. When training the model in Figure 2, the batch-size is 32, the number of raining epochs is set to be 3000, and the learning-rate is set to be 0.0000075. As the data is highly unbalanced, in one epoch of training, all the correct instances in the training data and the same size of incorrect instances sampled from the training data are used, and the incorrect instances are resampled every 50 epochs.

## 5.5 Feature Analysis

Data from the experiments support the hypothesis that the syntactic role of the target verb inside a sentence determines the occurrence and distribution of its complements. Consider Example (x-x+1) below:

Example 3 *像由**听**流行歌曲人们可以解除一些的压力。

Example 4 *通过人们**听**流行歌曲可以解除一些的压力。

Grammatical error is identified with 听'listen in both examples. In Example X, the verb acts as the head of a preposition phrase for 解除'release and requires the preposition to be 通过'by-way-of. In Example X+1, the subject 人们'people should be removed because the verb is the head of a preposition phrase and it requires that no subject should be present as an immediate complement.

This hypothesis is supported by t-SNE （T-distributed Stochastic Neighbor Embedding）with dependency types of the target verbs. Figure 12-13 are plots of t-SNE analysis of instances with *ccomp* and instances with *acl* as the CORE-DEP-in-CLS. In both cases it can be observed that there are clusters wherein the instances of correct usage are dominant, making discrimination possible.

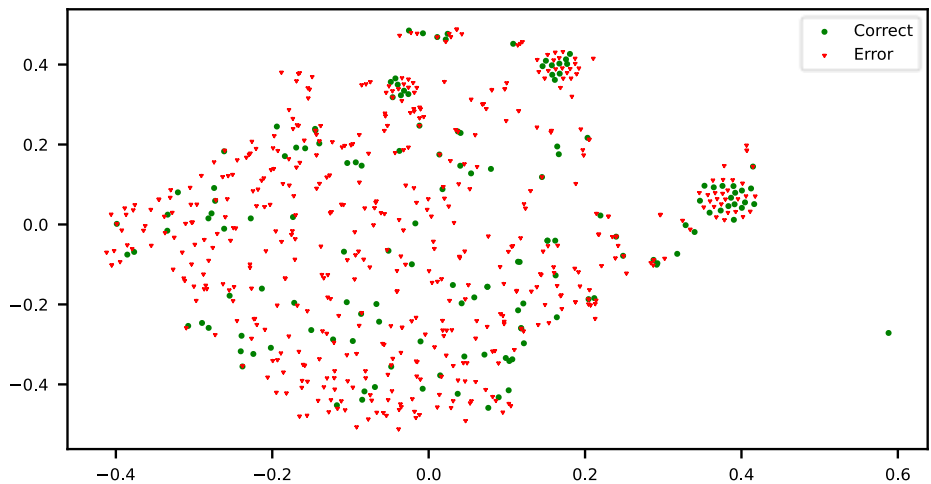

**Fig. 12 T-SNE analysis of core dependency type *ccomp* in *clause***

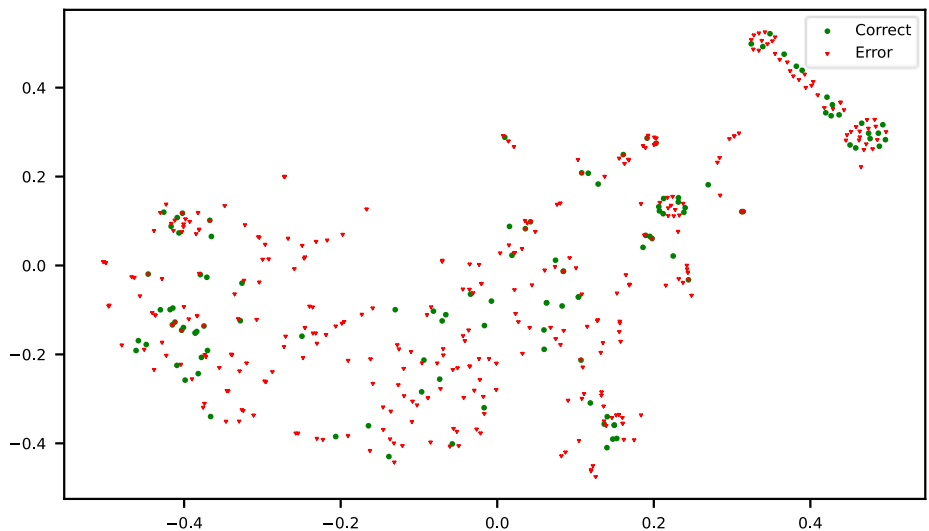

**Fig. 13 T-SNE analysis of core dependency type *acl* in *clause*.**

The above-mentioned hypothesis provides support for the MultiHeadAttention-based mechanism that integrates core-dep with dependency type in both $C_{top}$ and *clause* in the grammatical error identification model in Figure X. Given the hypothesis, the mechanism is designed to learn to discriminate correct instances from errors by observing unmatched dependency types.

**5.6 Evaluation against baseline**

To evaluate the performance of the current system, the performance of ChatGPT-4o is used as the baseline. The prompt used to obtain the performance of ChatGPT-4o via its web UI[8] is given in Example 4 below:

> Example 5
>
> 在下面的每一行中，找到 Target Verb:后的中文动词，Sentence:后的汉语句子，获取该中文动词的常规用法，并判断汉语句子中该动词的使用是否符合汉语语法，如果符合语法规则给出标记 0，不符合给出标记 1。判断完所有句子后按"序号->标记"列出结果。
>
> Sentence: 我要说的不是我们要回避，就是我们应该面对的问题，但是一定要保持"中庸"的道理，想人类的健康，还需要更多人的吃饱，而且环境保护也是不可忽略的问题。 Target Verb: 保持
>
> Sentence: 在那个时候学校放假大多数企业不要上班，大家都休息。 Target Verb: 休息
>
> Sentence: 不主动处理，谁替你能处理、解决呢？这样的人是没有进步的人，受埋怨的人。自己的责任换到别人的责任。人们明明知道将来的后果的时,绝对别依靠他人。
>
> Target Verb: 依靠

Table 11 gives the performance of ChatGPT-4o and the present study. The overall accuracy of ChatGPT-4o is lower than this study, but these two systems have their own advantages and disadvantages. In recognizing correct verb usage, ChatGPT-4o achieves better performance than this study. This can be explained in the general statement that rare verb usage is likely included in the huge volume data used to train ChatGPT-4o but is less likely available in the dataset used in this study. Meanwhile in recognizing verb usage errors, the present study outperforms ChatGPT-4o, supporting the hypothesis that an explicit knowledge of verb usage pattern can enhance the ability to identify improper verb usages.

---

[8] Available at https://chatgpt.com, accessed on March 20th, 2025.

**Table 11 Evaluation of grammar error detection against the base line**

| | Overall | Correct Verb Usage | | | Verb Usage Errors | | |
| --- | --- | --- | --- | --- | --- | --- | --- |
| | Accuracy | Precision | Recall | F-score | Precision | Recall | F-score |
| **ChatGPT-4o** | 0.564 | 0.298 | 0.579 | 0.394 | 0.804 | 0.561 | 0.661 |
| **Present study** | 0.612 | 0.288 | 0.516 | 0.370 | 0.823 | 0.639 | 0.720 |

# 6 Conclusions

The present study proposes a fully unsupervised approach to the construction of verb knowledge database, aimed at complementing LLMs by providing explicit and interpretable rules for application scenarios where explanation and interpretability are indispensable, and at reducing manual labor in constructing large scale knowledge database.

In order for the verb collostruction database to be functional as a source of event knowledge and applicable as a reference for prototypical syntactic-semantic patterns, it first develops a formal definition of the concept of verb collostruction with two design features—functional independence and graded typicality with negative evidence and then introduces the collostruction generating algorithm that uses clustering to ensure that the obtained collostructions to be functionally independent and prototypical. With a given verb, the BERT-based clustering is used to group input sentences to groups that each sharing a similar context, the DBSCAN-based clustering is used to cluster clauses that share similar syntactic patterns, and path traversal is used to generate subgraphs as collostructions from graphs obtained by merging clauses inside clusters.

Two methods are used to evaluate the verb collostructions obtained from the algorithm. By way of statistical analysis, it is demonstrated that the distribution of collostructions of a verb possesses the property of cascading self-similarity and that the collexemes inside each slot of the collostructions are semantically similar when measured with word embeddings. The evaluation with grammatical error correction also shows that an F-score of 0.612 is obtained by using collostructions as reference verb usage patterns to identify verb usage errors in L2 Chinese learners, which is higher than the baseline obtained with ChatGPT-4o. Both methods of evaluation demonstrate that the collostructions obtained with the algorithm meet the design features of functional independence and graded typicality.

As the proposed definition of verb collostruction and the algorithm used to generate collostruction for a given verb are not specific to Chinese, it is argued that the definition and the accompanying algorithm can be generalized to other languages and verb collostruction database can be fully automatically constructed for these languages. When verb collostruction databases are constructed for multiple languages, the design features of functional independence and graded typicality can be used for cross-language comparison of event characteristics.